\documentclass[10pt,twocolumn,letterpaper]{article}

\usepackage{wacv}
\usepackage{times}
\usepackage{epsfig}

\usepackage{graphicx}
\usepackage{amsmath}
\usepackage{amssymb}

\newcommand{\vfeat}{\mathbf{v}}
\newcommand{\afeat}{\mathbf{a}}

\newcommand{\hstate}{\mathbf{h}}
\newcommand{\acontext}{\mathbf{k}}
\pdfoutput=1

\wacvfinalcopy 

\def\wacvPaperID{236} 

\ifwacvfinal\fi
\begin{document}

\title{Action Classification and Highlighting in Videos}

\author{Atousa Torabi \\
Disney Research Pittsburgh\\
{\tt\small atousa.torabi@disneyresearch.com}
\and
Leonid Sigal \\
Disney Research Pittsburgh\\
{\tt\small lsigal@disneyresearch.com}
}

\maketitle
\ifwacvfinal\thispagestyle{empty}\fi

\begin{abstract}
   Inspired by recent advances in neural machine translation, that jointly align and translate using encoder-decoder networks equipped with attention, we propose an attention-based LSTM model for human activity recognition.
Our model jointly learns to classify actions and highlight frames associated with the action, by attending to salient visual information through a jointly learned soft-attention networks.
We explore attention informed by various forms of visual semantic features, including those encoding actions, objects and scenes.
We qualitatively show that soft-attention can learn to effectively attend to important objects and scene information correlated with specific human actions. Further, we show that, quantitatively, our attention-based LSTM outperforms the vanilla LSTM and CNN models used by state-of-the-art methods.  
On a large-scale youtube video dataset, ActivityNet \cite{activitynet}, our model outperforms competing methods in action classification.
\end{abstract}

\section{Introduction}

Activity recognition, classification and understanding are important problems in computer vision. The popularity of these problems, empowered by recent availability of large-scale on-line video resources (\eg, YouTube, Google, Bing) and datasets (\eg, ActivityNet \cite{activitynet}), stems from the vast number of applications that such technology will enable that include surveillance, video understanding, intelligent search and retrieval, automated metadata tagging and human-computer interactions. However, activity recognition has also proven to be an immensely challenging visual task as granularity, viewpoint, appearance and duration of activities can vary greatly and both spatial and temporal context appear to be critically important. 
\begin{figure}[ht]
\begin{center}
   \includegraphics[trim = 0mm 2mm 0mm 0mm, clip, width=0.47\textwidth]{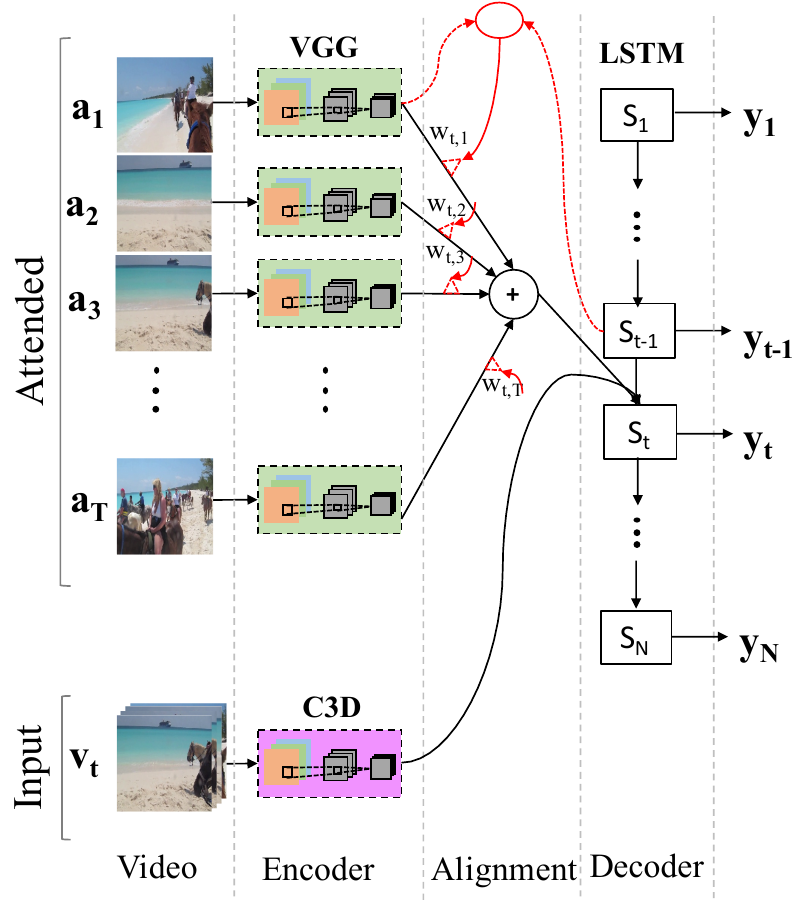}
\end{center}
   \caption{We use an encoder-decoder framework for action recognition. Our LSTM decoder is equipped with an attention/alignment model that allows it focus on salient context/attended data. Attended-data is encoded using VGG network \cite{Simonyan14c} and LSTM input-data is encoded using C3D network \cite{TranBFTP14}.}
\label{overview}
\label{fig:onecol}
\vspace{-0.2in}
\end{figure}

Neural network-based architectures, particularly Convolutional Neural Network (CNN), and end-to-end learning have recently illustrated great success in advancing state-of-the-art in variety of problems in computer vision, including image classification \cite{Krizhevsky_imagenetclassification,SzegedyLJSRAEVR14,Simonyan14c}, object detection \cite{girshick14CVPR,HoffmanGTDGDS14} and most recently action recognition \cite{SimonyanZ14,TranBFTP14}. However, such CNN models largely ignore the temporal progression of the action and typically learn frame-based or short 3-D spatio-temporal filters \cite{TranBFTP14} and aggregate information using simple maximum or average pooling across fixed-length segments \cite{SimonyanZ14,TranBFTP14} or the entire video. Very recently, in less than a handful of papers, Recurrent Neural Network (RNN), that have long history in speech recognition and language translation, have been re-discovered as effective means of more explicitly modeling temporal context in action recognition \cite{SrivastavaMS15,NgHVVMT15,WuWJYX15}. One significant limitation of RNN is inability to backpropagate error through long-range temporal interval (a problem known as “vanishing gradient” effect). A class of models that was specifically developed to be immune to such effect, through the use of simple memory-cell-like gates, was first introduced by Hochreiter and Schmidhuber \cite{Hochreiter1997} and termed Long Short-Term Memory (LSTM).  

LSTMs coupled with CNNs have many properties that make them appealing and notably appropriate for temporal perception tasks (hierarchical representation learning, temporal context with memory, possible end-to-end learning). However, such models effectively distill each frame into static learned representation, using one or more CNNs, and aggregate this(these) representation(s) across time using LSTM. One seemingly important property of human perception that is glaringly missing is {\em attention}, \ie, tendency not to process a whole video at once and rather attend selectively to parts of video that are most important for a given (\eg, activity) recognition task. We note that such attention mechanism is particularly important for videos that contain visual clutter. By attending, parts of the video that are most discriminative can be {\em highlighted} and used to better inform recognition model during learning and inference.  

In this paper, we explore the use of end-to-end encoder-decoder LSTM framework with the built-in attention mechanism. The most distinguishing feature of the proposed approach is that it does not attempt to utilize (encode) the entire input video for prediction of the action class (decoder). Instead it encodes a video into a temporal sequence of visual representations and chooses an adaptively wighted subset of that sequence for prediction. Each time model generates an action class prediction it soft-searches for a set of temporal positions within the video where most salient relevant information is concentrated. This is a much more flexible mechanism, as compared to, for example, earlier structured latent variable models \cite{raptisKS13}, that required much more restrictive conditional independence constraints, or fixed low-level attention schemes \cite{Mathe2012,Vig2012} which failed to significantly improve the performance or scale in practice. 

\vspace{0.05in}
\noindent
{\bf Contributions:} 
To the best of our knowledge, we are the first to show significant performance improvements that can come from temporal attention in action classification task. While in \cite{Serena972} attention-based LSTM is also proposed, the improvements due to attention network are shown to be minimal ($< 0.5$\%). In contrast, we show that by attending on i) much longer sequences and on ii) complementary semantic contextual features (\eg, objects, actions and scenes), much larger improvements in performance can be made (of up to 20\% or 8 percentage points). Further, we show that individual models, trained to attend on these different features, are themselves complementary and can be combined for additional performance gains. Conceptually, attention network allows our model to automatically learn to highlight salient spatial/spatio-temporal cues while integrating that information with LSTM model for time-step action classification. Qualitatively, we show that attention network learns to consistently attend to important semantics of the actions, despite the noise present in the input sequences, which adds to interpretability of the model.  Quantitatively, we show improved performance compared to the state-of-the-art on the large scale ActivityNet \cite{activitynet} dataset. 


\section{Related Work}

Recognition and understanding of human activities is a well studied problem in computer vision. As such, complete overview is beyond the scope of this paper and we refer reader to surveys \cite{Aggarwal2011,Poppe2010} for a more historical perspective. Here we focus on recent and most relevant approaches. 

\vspace{0.05in}
\noindent
{\bf Traditional action recognition:} Traditional action recognition approaches largely focused on global video representations that achieved good results on smaller and simpler datasets, \eg, KTH \cite{recognizinghuman}, HMDB51 \cite{hmdb51} and UCF101 \cite{ucf101}. Such approaches focused on pach-based local motion and appearance information in the form of Histogram of Oriented Gradients (HOG), Histogram of Optical Flow (HOF) \cite{Laptev2008}, Motion Boundary Histogram (MBH) \cite{Wang2010} or dense-trajectories \cite{wang.2011,wang.2013}. Classification was typically achieved by further aggregating these local representations across videos or using spatio-temporal pyramids \cite{Lazebnik2006} with Bag-of-Words(BoW) \cite{Laptev2008} or Fisher vector based encodings \cite{wang.2013} followed by traditional classifiers (\eg, SVMs). 


\vspace{0.05in}
\noindent
{\bf CNN-based approaches:} 
Availability of larger datasets and recent advances in representation learning for recognition, in the form of CNN, have led to the use of CNNs in action classification and understanding. The benefits of CNNs, \eg, over BoW, stem from their ability to learn relevant and discriminative representations, at a hierarchy of spatio-temporal granularities. Among the earliest, \cite{KarpathyCVPR14,ji.2013} used 3D CNN to encode few video frames,
showing that encoding multiple frames using raw pixels is marginally better than single-frame CNN models for video classification \cite{KarpathyCVPR14}. Simonyan \etal~\cite{SimonyanZ14} incorporated stack of 10-15 optical flow images, instead of raw pixels, in CNN model trained to encode motion information. Frame-level spatial information was learned using a separate CNN stream, and combination the two CNN models was shown to outperform traditional approaches. More recently, Tran \etal~\cite{TranBFTP14} learned generic video feature, in the form of 3D CNN, and applied it to variety of video understanding tasks. 
To incorporate some longer-term temporal information, Zha \etal~\cite{ZhaLASS15} experimented with different spatial and temporal pooling over CNN features, applying the resulting model to TRECVID multimedia event detection challenge; similarly, Ng \etal~\cite{NgHVVMT15} study variety of CNN pooling strategies that process up to 120 frames and obtain impressive results on Sport-1M video classification.


\vspace{0.05in}
\noindent
{\bf Temporal models and LSTM:} 
The core idea that temporal progression of an activity is important, has a long history in the field. A variety of early models have been utilized for modeling of temporal context, including Hidden semi-Markov Models (HSMM) \cite{Duong2005}, CRFs \cite{Sminchisescu2005}, and finite-state-machines \cite{Ikizler2007}. Most recently, RNNs have also been explored; LSTM, a form of RNN, has recently become particularly popular due to its ability to deal with vanishing and exploding gradients during training. RNN/LSTM have a number of notable benefits with respect to, for example, HMMs or CRFs, including fewer conditional independence assumptions and possible end-to-end training when combined with CNN. Several recent works incorporate spatial/optical-flow CNN features with vanilla LSTM models for global temporal modeling of videos \cite{SrivastavaMS15,DonahueHGRVSD14,NgHVVMT15,WuWJYX15,VenugopalanRDMD15}. These works showed improvements with respect to action recognition \cite{DonahueHGRVSD14,Lazebnik2006,WuWJYX15} or video description tasks \cite{yao2015capgenvid,VenugopalanRDMD15}. Similar to these methods, we also incorporate CNN features with LSTM for action recognition. However, we use a new variant of LSTM model equipped with an attention network that allows it to focus and highlight discriminative frames and use object and scene features as context.

\vspace{0.05in}
\noindent
{\bf Temporal and spatial attention:} 
There have been previous attempts to ask whether information from all frames is equally important for video recognition/understanding \cite{Schindler2008}. To this end a number of keyframe-based representations have been proposed. For example, \cite{Zhao2008} rank all frames based on holistic information theoretic measure; \cite{Liu2012} rely on spatio-temporal localization and AdaBoost to select keyframes; \cite{raptisKS13} automatically select compact keyframe representation using a latent structured SVM formulation. In complement, low-level spatial attention models, trained from eye-gaze in video, have been studied in \cite{Mathe2012,Vig2012}, with relatively limited success. More recently, Shapovalova \etal \cite{NIPS2013_5197} use eye-gaze information as weak-supervision for action localization. Our model also has an attention mechanism, but compared with \cite{raptisKS13} or \cite{NIPS2013_5197} that suffer from complexities imposed by discrete latent variables, avoids scalability issues by employing soft-attention coupled with LSTM. Our {\em dynamic} soft-attention mechanism is also in stark contrast with static low-level attention models of \cite{Mathe2012} and \cite{Vig2012}. 

\vspace{0.05in}
\noindent
{\bf LSTM models with attention:} 
Models that combine neural network architectures with attention are relatively recent (perhaps the earliest mention is in the context of Boltzmann Machine \cite{Larochelle2010}). 
In vision, Xu \etal~\cite{XuBKCCSZB15} use spatial soft-attention that learns to fix gaze on salient image parts while generating description for images. Yao \etal~\cite{yao2015capgenvid} also use soft-attention mechanism on video frames for video description. Perhaps in the closest work to ours, Yeung \etal~\cite{Serena972} propose a multiLSTM for action detection, where as input-data their model aggregates few frames CNN features using soft-attention and generates multiple outputs at each time step corresponding to multiple frames. The results show that adding soft-attention for weighted averaging of few input frames only marginally ($< 0.5$\%) improves the performance over uniform average. In contrast, we apply soft-attention mechanism on temporally longer contextual information (\eg, object, scene or complementary action features) rather than LSTM inputs. We show that in this way our jointly trained model can better focus on, or highlight, important frames, which substantially improves action classification performance and gives some semantic interpretation for results.


\section{Approach}
\label{approach}
 In our model, activity recognition is defined as sequential  problem where inputs are video features, $\vfeat_t$, over time and goal is to predict a single label from a fixed action vocabulary $\left\langle \vfeat_{1}, \vfeat_{2} ..., \vfeat_{t} \right\rangle \rightarrow y$. 
 We make use of the encoder-decoder framework, where the input video features are encoded using CNN models and the prediction is done using LSTM as a decoder. Figure \ref{overview} shows the overview. 
 
 In particular, we build on attention-based LSTM \cite{BahdanauCB14} for machine translation, and propose a model that takes two separate sets of input video features: video-level spatio-temporal features that we call {\em Input-data} and frame-level semantic image features (\eg, encoding objects and scenes) that we call {\em Attended-data}. In our model, Input-data and Attended-data are encoded using different CNN models and are described in detail in Sections \ref{input-data} and \ref{attended-data} respectively. We use soft attention mechanism that learns to highlight (or weight) frames with relevant visual semantics (\ie, objects or scenes important for an action label prediction)\footnote{In the context of attention-based LSTM models this is often called {\em soft alignement}, where in this case alignment is between the output action label and the input frames.}. 
 We describe our soft attention model for action recognition task in Section \ref{alignment} and the final LSTM decoder in Section \ref{lstm-decoder}.           


\subsection{Input-data: CNN Encoder}
\label{input-data}
We assume the input-data to our encoder is a fixed temporal-length window of $F$ frames.  
We extract video features from this window using spatio-temporal CNN implemented in Caffe \cite{jia2014caffe}. As a result, input window is represented using a matrix
\begin{equation}
\label{equ:input}
\textbf{V} = \left\{\vfeat_{1}, ..., \vfeat_{N} \right\} \in \mathbb{R}^{N \times d_\vfeat} \end{equation}
of $N$ video feature vectors, where each vector $\vfeat_{i} \in \mathbb{R}^{d_\vfeat}$ is a $d_\vfeat$-dimensional representation corresponding to a chunk of $F / N$ frames. Note, that $N$ will also correspond to the length of the LSTM decoder sequence; \ie, for every video feature vector we will get an activity label prediction.

In particular, we make use of C3D network pre-trained on Sport-1-million dataset and fine-tuned it on trimmed ActivityNet videos for activityNet experiments.
C3D network, proposed in \cite{TranBFTP14}, has 8 convolutional layers and 2 fully connected layers, followed by a softmax output layer. All 3D convolution kernels are $3 \times 3 \times 3$ with stride 1 in both spatial and temporal dimensions. 
We extract first Fully-Connected layer (FC6) as a feature vector representation, therefore $d_\vfeat = 4096$. 

\subsection{Attended-data: CNN Encoder}
\label{attended-data}
 For attended-data, we extract video features using spatial CNN in Caffe \cite{jia2014caffe}. As a result, attended-data is represented using a matrix,
\begin{equation}
\label{equ:attended}
\textbf{A} = \left\{\afeat_{1}, ..., \afeat_{T} \right\} , \afeat_{i} \in \mathbb{R}^{T \times d_\afeat}, \end{equation}
of $T$ video feature vectors where each vector $\afeat_{i} \in \mathbb{R}^{d_\afeat}$ is a $d_\afeat$-dimensional representation corresponding to a single frame. In our model each attended frame is sampled every $F / N$ video frames,  therefore $T = N$. We make use of VGG-19 network \cite{Simonyan14c} trained for three task of object, action, and scene recognition: 
\vspace{-0.05in}
\begin{itemize}
\item $VGG_{obj}$: VGG trained, on ImageNet, for $1000$ class object classification \cite{Simonyan14c}; 
\item $VGG_{act}$: VGG-19 network fine-tuned on ActivityNet data for action recognition of $203$ action classes; 
\item $VGG_{sce}$: VGG-19 network fine-tuned on MIT-Scenes for scene recognition of $205$ scene classes \cite{NIPS2014_5349}. 
\end{itemize}
\vspace{-0.05in}
\noindent
VGG-19 proposed in \cite{Simonyan14c}, has 16 convolutinal layers and 3 fully connected layers, followed by a softmax output layer. For some experiments we extract first Fully-Connected layer (FC6), therefore $d_\afeat = 4096$, and for others we extract last Fully-Connected layer (FC8) as a frame representation. For $VGG_{obj}$ FC8 layer $d_\afeat = 1000$, for $VGG_{act}$ FC8 layer $d_\afeat = 203$, and for $VGG_{sce}$ FC8 layer $d_\afeat = 205$.

In general, size of attended-data $T$ could be different from size of input-data $N$ ($T \ne N$). For example, one may want to attend to objects/scenes/actions at a coarser or finer temporal granularity compared to the granularity of label predictions. In experiments, we also consider effect of finer grained attention, \eg, sampled every $F/(2N)$ such that $T=2N$. We also evaluated significance of attended-data size, comparing $T=12$ and $T=24$. We only observed a negligible improvement using bigger attended-data size.

\subsection{Action Attention and Highlighting Model}
\label{alignment}

Attention model is parametrized as a feedforward neural network that is jointly trained with LSTM decoder (see Section \ref{lstm-decoder}). In other words, the gradients of LSTM loss function are backpropagated not only through LSTM network but also through 
soft-attention network that allows joint training of both 
networks. This model has been originally proposed by Bahdanau \etal \cite{BahdanauCB14} for machine translation. We adopted this model for  
action highlighting with attended-data that we described in the previous section. 
Attention model generates a score for each attended frame $\afeat_{i}$ at time step $t$. This score represents how well a sequentially modeled activity video up to time $t-1$ and attended $\afeat_{i}$ are semantically matched. In other words, matching score $m_{t,i}$ measures the relevance of $\afeat_{i}$ and LSTM decoder hidden state at time $t-1$ and is defined as    

\begin{equation}
\label{equ:energy}
m_{t,i} = \Phi( \hstate_{t-1}, \afeat_{i} ) ,
\end{equation}

\noindent
where $\hstate_{t-1}$ is LSTM decoder hidden state at time $t-1$ that contains all information related to a temporally modeled video sample up to time step $t-1$. For 
model $\Phi$, every state $\hstate_{t-1}$ is summed with every attended-data $\afeat_{i}$ to obtain a matching-vectors. Then matching-vector is transformed into a single number named matching-score $m_{t,i}$. Note, the 
matching-vector dimension is one of the network parameters and, in general, could have different value than LSTM hidden-state. Finally, attention weights are computed by normalizing matching-scores in a softmax-like function. Attention weight $w_{t,i}$ for attended frame $i$ at time $t$ is computed by: 
\label{alignment}
\begin{equation}
\label{equ:weight}
w_{t,i} = \frac{exp(m_{t,i})}{\sum^{T}_{j=1}exp(m_{t,j})}.
\end{equation}
%
A higher attention weight reflects more saliency attributed to a specific attended frame $i$ with respect to the specific task (action classification).    

\vspace{0.05in}
\noindent
{\bf Action Attention Context:}
In order to integrate attended-data with LSTM decoder, we compute the weighted average of the attended-data using the attention weights obtained above. We term this procedure - {\em action attention context}: 
\begin{equation}
\label{equ:context}
\acontext_{t}(A)  = \sum^{T}_{i=1}w_{t,i} \afeat_i.
\end{equation}

\subsection{Attention-Based LSTM Decoder}
\label{lstm-decoder}
Long Short-Term Memory (LSTM) is a specific type of a recurrent neural network that recently has been widely used in various domain for sequential modeling, due to its capabilities to cope with "vanishing gradients" problem and its "internal memory" property. In high-level definition, our attention-based LSTM decoder $\psi$ sequentially updates LSTM state $\hstate_{t}$ and internal memory $\mathbf{c}_{t}$ given previous state $\hstate_{t-1}$, current input-data $\vfeat_{t}$, and action attention context vector $\acontext_{t}(A)$. Also, simultaneously, it predicts action label $\mathbf{y}_{t}$ at time $t$. In abstract, our LSTM decoder is defined as following: 

\begin{equation}
\label{equ:RNN}
\begin{bmatrix} \hstate_{t} \\ \mathbf{y}_{t} \\ \mathbf{c}_{t} \end{bmatrix} = \psi( \hstate_{t-1}, \vfeat_{t}, \acontext_{t}(A) ).
\end{equation}

In order to fuse input-data and attended-data, the information from {\em action attention context} are distributed across the input-data $\vfeat_t$. We used the version of LSTM in \cite{Graves13}, where $\hstate_{t}$ is updated by the following intermediate functions:
\begin{eqnarray}
\label{equ:LSTM}
\mathbf{i}_{t} & = &\sigma(W_{i} \vfeat_{t} + U_{i} \hstate_{t-1} + \mathbf{b}_{i}) \\
\mathbf{f}_{t} & = & \sigma(W_{f} \vfeat_{t} + U_{f} \hstate_{t-1} + \mathbf{b}_{f}) \\
\mathbf{o}_{t} & = & \sigma(W_{o} \vfeat_{t} + U_{o} \hstate_{t-1} + \mathbf{b}_{o}) \\
\mathbf{g}_{t} & = & \tanh(W_{c} \vfeat_{t} + U_{c} \hstate_{t-1} + \mathbf{b}_{c})\\
\mathbf{c}_{t} & = & \mathbf{f}_{t} \mathbf{c}_{t-1} + \mathbf{i}_{t} \mathbf{g}_{t} \\ 
\hstate_{t} & = &  \mathbf{o}_{t} \tanh(\mathbf{c}_{t}) ,
\end{eqnarray}

\noindent
where $\mathbf{i}_t$, $\mathbf{f}_t$, $\mathbf{o}_t$, and $\mathbf{c}_t$ are respectively input gate, forget gate, output gate, and cell gate (\ie, internal memory) at time $t$. In brief $\mathbf{i}_t$, $\mathbf{f}_t$, and $\mathbf{o}_t$ control influence of current input $\mathbf{g}_t$ and previous memory state $\mathbf{c}_{t-1}$ on generating new hidden state $\hstate_{t}$. The terms $W_i, W_f, W_o, W_c, U_i, U_f, U_o, U_c$ encode the various weight matrices and $\mathbf{b}_i, \mathbf{b}_f,\mathbf{b}_o,\mathbf{b}_c$ corresponding biases. 
The prediction $\mathbf{y}_{t}$ is computed by linear transform on hidden state $\hstate_{t}$ followed by a softmax:
\begin{equation}
\label{equ:RNN}
\mathbf{y}_{t} = softmax(W_{y} \hstate_{t} + \mathbf{b}_{y}),
\end{equation}

\noindent
where $\mathbf{y}_{t} \in \mathbb{R}^C$ is probability distribution over fixed number of $C$ classes. 

\vspace{0.05in}
\noindent
{\bf Action classification: }
We average the scores of $N$ time-steps of LSTM for each video chunk of $F$ frames (more details about classification are in Section~\ref{details}).

\section{Experiments}
\label{dataset}
We use ActivityNet dataset. ActivityNet \cite{activitynet} is a recent large-scale fine-graned human activity dataset with diverse $203$ classes organized in an ontology. ActivityNet includes $190$ Youtube video samples per class, on average, and video samples average $1382.3$ frames (frame rate is not identical for all the videos, however, for most videos it is $29$ frame per second). Samples correspond to trimmed activities from longer videos. An action label is assigned to each trimmed video clip. The trimmed data and labels are provided as part of standard activity classification task defined in \cite{activitynet}. However, the trimmed videos are very noisy and contain various visual concepts which are not directly related to the action label. 
The dataset includes three splits corresponding to training, validation, and test sets. The test set is not available (withheld by the authors). In our experiment we use video samples with minimum 192 frames (about $90\%$ of videos). To this end, we split the original 25,124 training split into 22,624 training samples and 2,500 validations samples for training our algorithm. We used ActivityNet validation split, defined in \cite{activitynet}, consisting of 8,403 videos, as our test set. 

\subsection{Implementation Details}
\label{details}
\vspace{-0.01in}
We implemented our attention-enabled LSTM decoder in Theano-based framework named BLOCKS \cite{Bastien-Theano-2012,MerrienboerBDSW15}. Our data pipeline has been implemented using FUEL framework \cite{MerrienboerBDSW15}. Our LSTM models have $N=12$ time-steps and we set hidden-state dimension to $512$. The matching-vector dimension is also $512$. Each video sample has fixed size of $F = 192$ frames. The input-data is encoded by C3D network every $F/N$ frames which here is $16$ frames (see Section~\ref{input-data} for details). The choice of $16$ frame inputs for C3D network was made in light of experimental findings in \cite{TranBFTP14}. Attended-data is encoded by VGG networks applied to frames sampled every $16$ frames, from the same video of length $192$ (see Section~\ref{attended-data} for details). We added a mlp fully-connected layer before LSTM. This layer has a segmoid non-linearity applied on input-data without changing input-data dimension. For ActivityNet classification, we use softmax loss function.

\vspace{0.05in}
\noindent
{\bf Training:}
We train with stochastic gradient descent and apply dropout with probability $0.5$ to all LSTM non-recurrent connections and fully-connected layer. 
 
For ActivityNet, we use a fixed $0.01$ learning rate for $2,512$ iterations on minibatches of size $200$. After that, the learning rate is decreased every $125$ iterations by a fixed ratio. We also used a fixed $0.9$ momentum and use early stopping strategy on validation set loss. We augment the dataset by adding horizontally-flipped frames to training data. We perform an on-the-fly temporal cropping of training videos which means that at each iteration we randomly crop a fixed-size video chunk of $192$ frames from each video sample in the mini-batch.

\vspace{0.05in}
\noindent
{\bf Testing:}
For each test video, we randomly crop $40$ video chunks of $192$ frames and $40$ video chunks of $192$ horizontally flipped frames. The class scores for a test video clip are then computed by averaging the scores across all $80$ chunks. 
\begin{figure}[!t]
\begin{center}
   \includegraphics[width=0.475\textwidth]{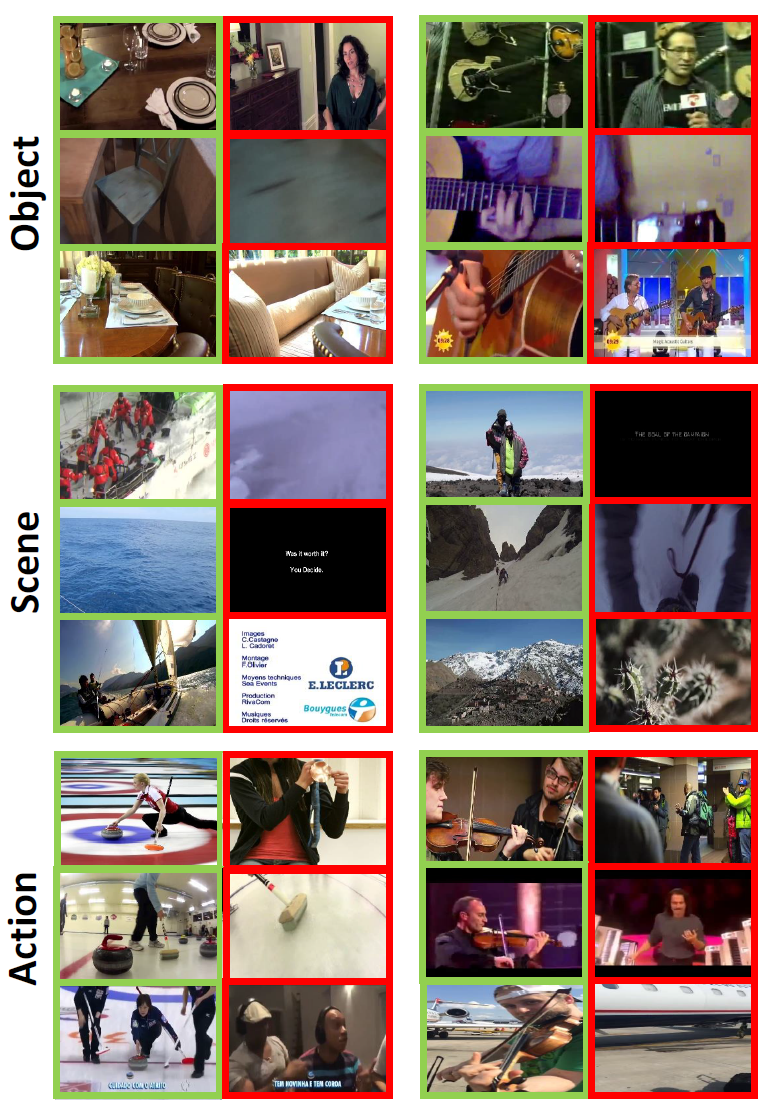}
\end{center}
   \caption{Three examples show frames with highest (green) and lowest (red) attention weights for each of the classes (left-to-right and top-to-bottom): {\em setting the table}, {\em playing guitar}, {\em sailing}, {\em mountain climbing}, {\em curling}, and {\em playing violin}. For {\em setting table} and {\em playing guitar} examples attention is on $objects$ context ($a$LSTM-$VGG_{obj}$ (fc8)), for {\em sailing} and {\em climbing mountain} examples attention is on $scenes$ context ($a$LSTM-$VGG_{sce}$ (fc8)), and for {\em curling} and {\em playing violin} examples attention is on $action$ context ($a$LSTM-$VGG_{act}$ (fc8)).}
\label{scene-object}
\label{fig:onecol}
\end{figure}

\vspace{0.05in}
\noindent
{\bf Evaluation Metrics:}
For ActivityNet, we compute the average accuracy over all the trimmed validation samples (used as test set; see above). We also report mean Average Precision (mAP) over all trimmed validation samples and over all $203$ classes.
%
%
 \begin{figure*}[!t]
\begin{center}
    \includegraphics[width=0.7\textwidth]{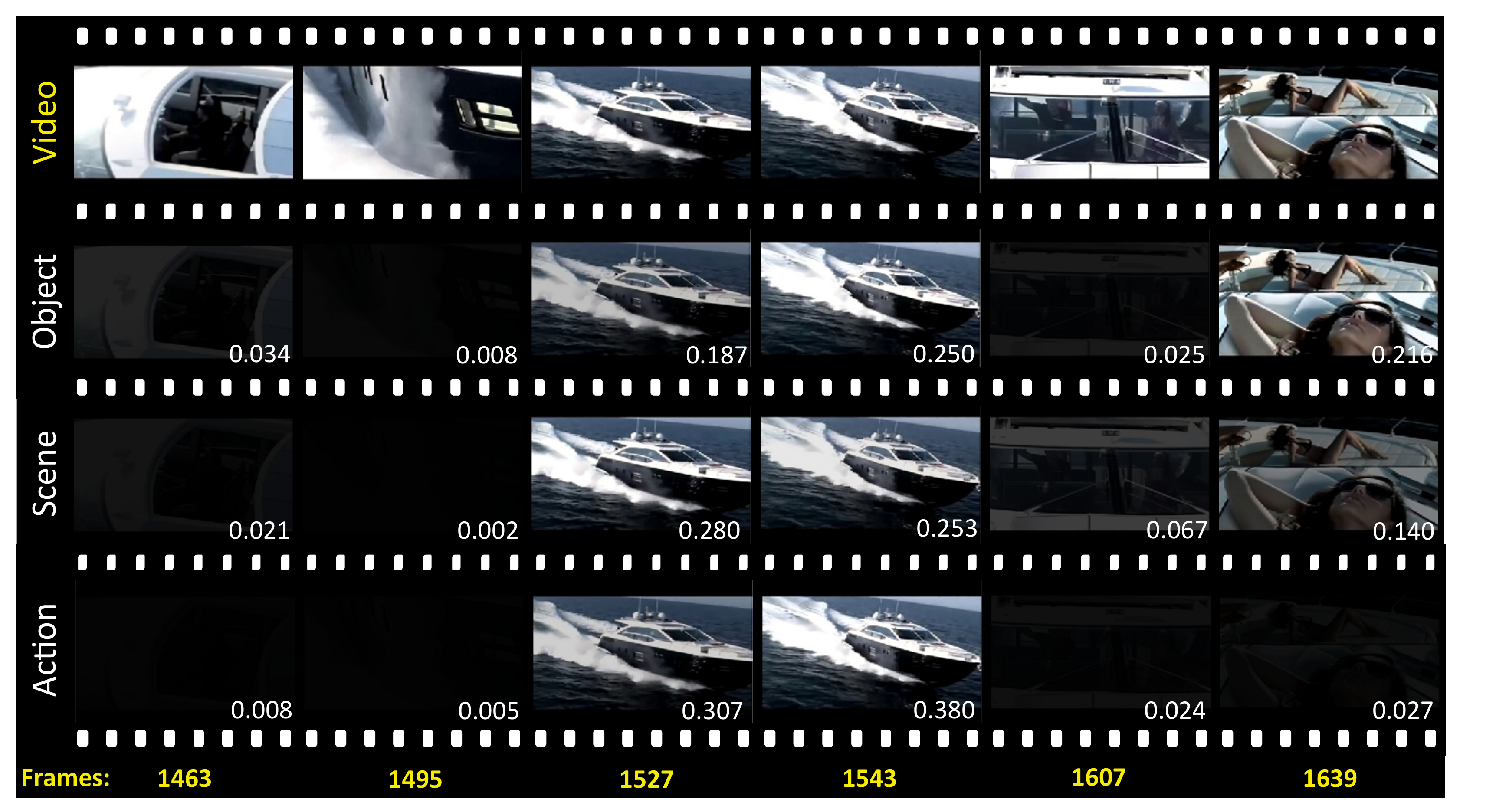} \\
    (a) Action: {\em Yatching} \\
    \includegraphics[width=0.7\textwidth]{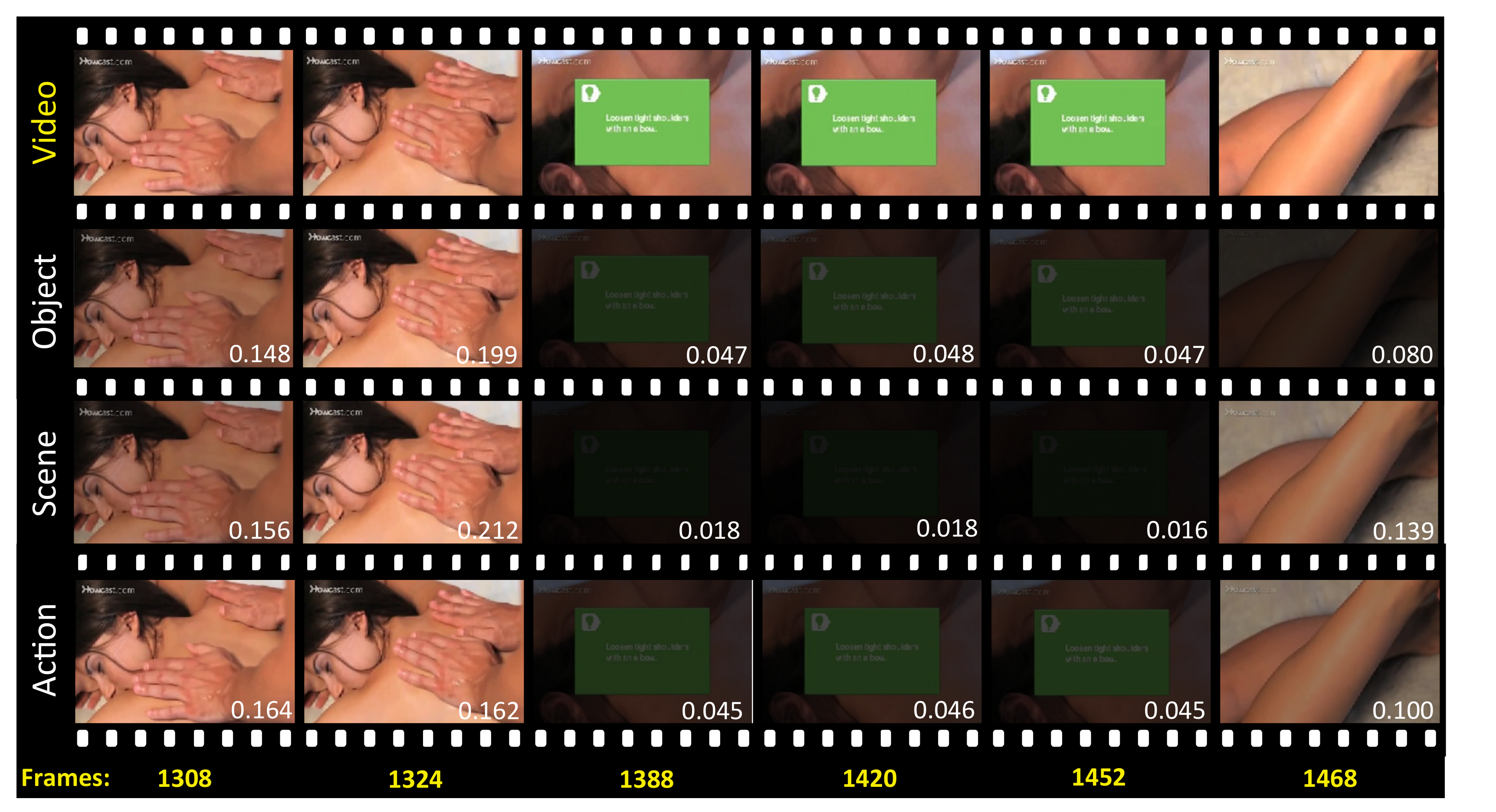} \\
    (b) Action: {\em Having Massage}    
\end{center}
\vspace{-0.1in}
   \caption{Visualization of our $a$LSTM-$VGG$ (fc8) models attention weights, attending on object, scene, and action for ActivityNet dataset. The lighter video frames have higher attention weight and darker frames represent lower attention weight.}
\label{attention-visualization}   
\label{fig:long}
\vspace{-0.1in}
\end{figure*}

 
\subsection{Qualitative Results}
In order to show how well our attention networks align action with contextual data, we present multiple visualizations. Figure \ref{scene-object} shows the video frames with the highest (green rectangle) and lowest (red rectangle) attention weights for $3$ different samples of each class. The top row shows examples for class {\em setting the table} and {\em playing guitar} respectively. For these examples contextual data (attended-data) are extracted from $VGG_{obj}$. You can see that for first example, images containing {\em table}, {\em chair}, {\em cup}, {\em dish}, {\em glass}, {\em candle} and {\em napkin} have maximum attention weights and image with {\em big couch}, very zoomed image, or image of a {\em woman sitting in a bedroom} have the minimum attention weights. In next example, the alignment model focus on {\em guitar} images and occluded images with text or an interviewer picture have the minimum attention weight for class {\em playing guitar}. Middle row shows $3$ video examples for class {\em sailing} and {\em mountain climbing} respectively. For these examples the attended features are extracted using $VGG_{sce}$. In the first example, images related to the scene including {\em sea}, {\em sky} and {\em sailing boat} have the highest attention weights and images that contain no scene salient features, related to sailing, have the minimum attention weights. Also in the next example, scenes showing {\em mountain}, {\em climbers}, {\em sky}, and {\em snow} have the highest attention weights and close-up images or text, with no features related to {\em mountain climbing}, have minimum attention weights. Finally, last row shows $3$ video examples for class {\em curling} and {\em playing guitar}. For these examples the attended features are extracted using $VGG_{act}$. As one can see, most images with maximum attention weight mainly contain {\em curling} and {\em playing guitar}. Besides showing that attention concentrates on relevant objects, actions and scenes, the figure also shows that there is reasonable alignment/visual consistency across the attended frames for each action.  

Figure \ref{attention-visualization} shows $2$ video samples for $2$ action classes. In this figure we show $6$ sampled frames out of $12$ from original video that are passed to $a$LSTM for context; weighted video frames for attention-based model focus on salient object features, salient scene features and salient action features respectively. For each example, we picked $6$ frame samples that show variety of attention weights. The darker images represent lower attention weights, therefore less significant features, and the lighter images have higher attention weights, therefore containing more salient aspects related to corresponding action class. For all examples attention mechanism mainly focus on salient and discriminative video frames. However, interestingly, they are some variations between what different models attend to. For instance, in {\em having massage} example, for $a\text{LSTM}$ model attending on scene, images containing big text box have minimal weight, since those images have no salient features related to natural scenes. For $a$LSTM model attending on object, in {\em yatching} example, last image has relatively high attention weight. In fact, if you search for {\em yatching} on Google, the first category that comes up is life style and top example photos are images of people in swimwear on boats. 

 \begin{figure*}[!t]
\begin{center}
   \includegraphics[width=0.8\textwidth]{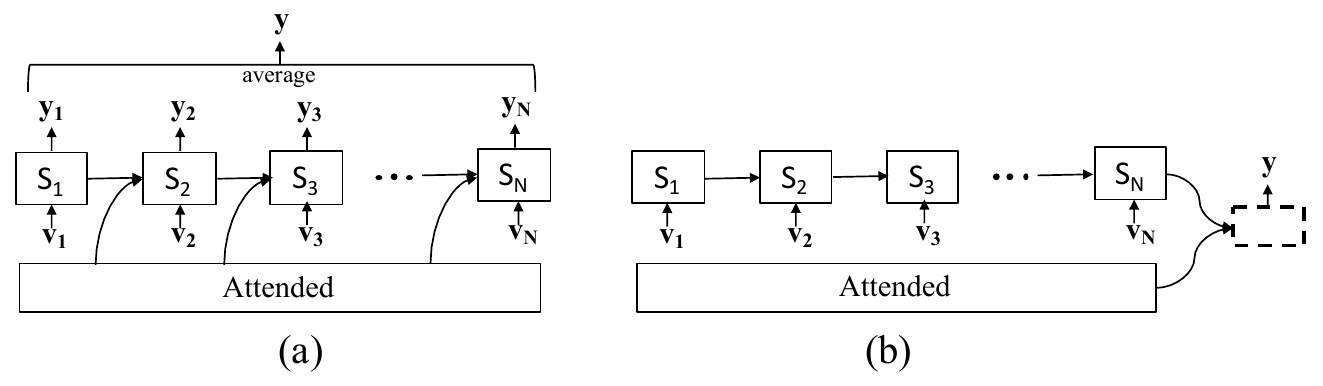}
\end{center}
    \vspace{-0.27in}
   \caption{Two variations of our attention-enabled LSTM architectures.}
\label{fig:figVar}
\vspace{-0.15in}
\end{figure*}

\begin{table}[!t]
\begin{center}
\caption{ActivityNet: Our model action recognition results.}
\begin{tabular}{|lr|c|c|c|c|c|c|c|}
\hline
\multicolumn{4}{|c|}{{\bf Model:} CNN} \\
\multicolumn{2}{|c|}{Features} & AC & mAP\\
\hline
$C3D$ & (fc8)   & 40.9\% & 40.0\% \\
\hline
\hline
\multicolumn{4}{|c|}{{\bf Model:} LSTM} \\
\multicolumn{2}{|c|}{Features} & AC & mAP \\
\hline
LSTM-$C3D$ & (fc6)  & 40.3\% & 40.6\% \\
\hline
$a$LSTM-$C3D$ ~~~~~~~~& (fc8)  & 42.7\% & 42.5\%\\
\hline
$a\text{LSTM-}VGG_{obj}$ & (fc6) & 45.7\% & 46.2\%\\
\cline{1-4}
 $a$LSTM-$VGG_{obj}$
    & (fc8) & 47.4\% & 48.0\%  \\
\hline
$a\text{LSTM-}VGG_{sce}$ & (fc6) & 44.0\% & 43.9\%\\
\cline{1-4}

$a\text{LSTM-}VGG_{sce}$& (fc8) & 45.3\% & 44.9\%\\
\hline
$a\text{LSTM-}VGG_{act}$ & (fc6) & 47.0\% & 47.3\%\\
\cline{1-4}
  $a\text{LSTM-}VGG_{act}$  & (fc8) & \textbf{48.1}\% & \textbf{48.6}\%\\
\hline
\end{tabular}

\label{tab:c3d}
\end{center}
\end{table}

 \begin{figure*}[!t]
\begin{center}
   \includegraphics[width=\textwidth]{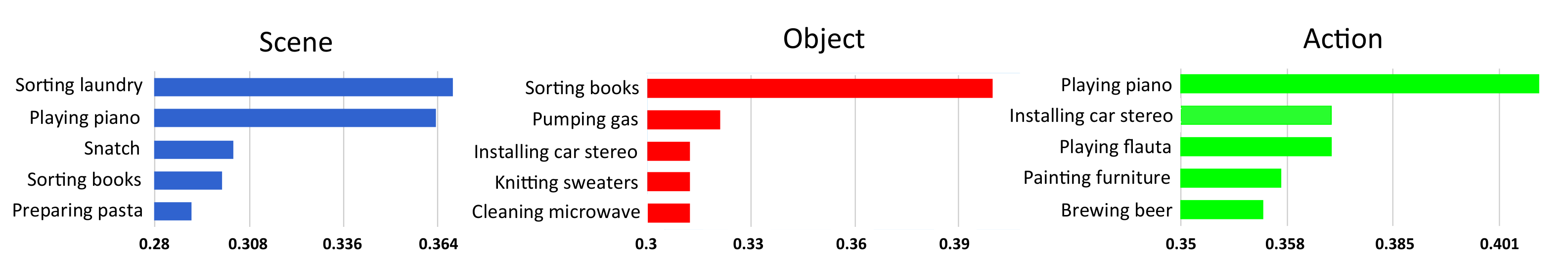}
\end{center}
\vspace{-0.25in}
   \caption{For each of the scene, action, object attended models the top 5 classes of ActivityNet on which improvement is achieved with respect to vanilla LSTM are shown. Note that on some classes improvement is as much as 40\% in classification.}
\label{bestclass}
\label{fig:onecol}
\vspace{-0.2in}
\end{figure*}

\subsection{Comparative Study of Our Model Variants}
\vspace{-0.1in}
Table \ref{tab:c3d} shows average accuracy (acc) and mean average precision over all classes (mAP) for the ActivityNet. 
$C3D (fc8)$ shows results for 3D CNN model softmax classification directly. LSTM-$C3D$ (FC6) is vanilla LSTM model using $C3D$ model fc6 layer representation as input-data. $a$LSTM stands for our attention-based LSTM model. We have evaluated our model for all different possible variations of attended-data using VGG model (note that for all three variations input-data is $C3D (fc8)$ features): 1) $a\text{LSTM}-VGG_{obj}$, attends on objects ($VGG_{obj}$), 2) $a\text{LSTM}-VGG_{sce}$, attends on scenes ($VGG_{sce}$), and 
3) $a\text{LSTM}-VGG_{act}$ attends on actions ($VGG_{act}$).
For each of these three models we tried both fc6 and fc8 layer representations. 
A few observations can be made based on Table \ref{tab:c3d}. First, LSTM on top of $C3D$ features has minor improvement over $C3D$ CNN for ActivityNet. All $a$LSTM models outperform vanilla LSTM and CNN models. Specifically, $a$LSTM on $C3D$ improves the performance by approximately $2$ percentage points for ActivityNet.
%
Further for ActivityNet, attending on (contextual) complementary trained VGG19 models for scene, object, and action improves the performance by a large margin, between $4-5$ percentage points for the same fc6 features; using fc8 features is uniformly better, even further improving performance.

Among individual models, the best performance is achieved by $a$LSTM-$VGG_{act}$ (fc8) model for ActivityNet (likely because $VGG_{act}$ has been fine-tuned for ActivityNet), which is consistent with recent results that show importance of objects \cite{Jian2015}. 

Figure \ref{fig:figVar} shows two possible variants of integrating attention with LSTM for action classification. In the first variant, Figure \ref{fig:figVar} (a), attention scores are computed at every time step $t$ and attended data is distributed into LSTM state $S_{t-1}$ to compute state $S_{t}$. Also, class scores $y_t$ are computed at every time step and final score $y$ is the average of all $y_t$s. This is the model we focused on in the approach section. 
The second variant, Figure \ref{fig:figVar} (b), has a somewhat different architecture. In particular, attention score is only computed and integrated into the last LSTM state $S_{N}$. 
The classification score $y$ is then computed directly (once) at the end. We compare both architectures, using identical setup, and we observed architecture (a) that we used for our experiment has better performance.

\vspace{-0.05in}
\subsection{Comparison With Competitive Models}

Table \ref{tab:comparison} shows comparison of our model to state-of-the-art on both AcivityNet. In \cite{activitynet}, $DF$ stand for Deep features, $MF$ stands for motion features, and $SF$ stands for spatial features (see \cite{activitynet} for more details). $SVM-DF$ (fc8) is one-vs-all SVM model over $DF$ fc8 layer representation described in \cite{activitynet}. 

For comparison, we have combined different variations of our models described above. Model combination has been implemented by averaging the softmax output scores over multiple $a$LSTM models. For example, $a$LSTM-$VGG_{act+obj}$ stands for combining $a$LSTM-$VGG_{act}$ and $a$LSTM-$VGG_{obj}$ and so on. Also $SVM_{MF+DF}$ stands for $SVM$ model over $MF$ and $DF$ feature concatenation described in \cite{activitynet}.
Table \ref{tab:comparison} shows that for ActivityNet our combined models improve both AC and mAPs up to $\sim 11.5\%$ (5.6 percentage points) compared to our best individual model $a$LSTM-$VGG_{act}$ (fc8). This reveals that variations of $a$LSTM that attend on $object$, $scene$, and $action$ contextual data to some extends have complementary information, where the combined model $a$LSTM-$VGG_{act+obj+sce}$ (all) has the best performance. 

\begin{table}[!t]
\begin{center}
\caption{ActivityNet: Comparative results.}
\begin{tabular}{|lr|c|c|c|}
\hline
\multicolumn{4}{|c|}{{\bf Model:} CNN} \\
\multicolumn{2}{|c|}{Features} & AC & mAP \\ 
\hline
CNN-$C3D$ & (fc8)   & 40.9\% & 40.0\%\\
\hline
SVM-$DF$ \cite{activitynet} & (fc8)  & - & 38.1\% \\ 
\hline
\hline
\multicolumn{4}{|c|}{{\bf Model:} Combined} \\
\multicolumn{2}{|c|}{Features} & AC & mAP \\ 
\hline
SVM$_{MF+DF}$ \cite{activitynet} &  & - & \multicolumn{2}{|c|}{49.5\%} \\
\hline
SVM$_{MF+SF}$ \cite{activitynet} &  & - & 48.9\% \\ 
\hline
SVM$_{MF+DF+SF}$ \cite{activitynet} ~& & - & 50.5\% \\ 
\hline
\small $a\text{LSTM-}VGG_{act+obj}$ & (fc6) & 50.0\% & 51.0\%\\
\cline{1-4}
\small $a\text{LSTM-}VGG_{act+obj}$ & (fc8) & 50.7\% & 51.4\%\\
\hline
\small $a\text{LSTM-}VGG_{act+sce}$ & (fc6) & 50.4\% & 50.7\%\\
\cline{1-4}
 \small $a\text{LSTM-}VGG_{act+sce}$    & (fc8) & 50.7\% & 51.2\%\\
\hline
\small $a\text{LSTM-}VGG_{act+obj+sce}$ & (fc6) & 51.4\% & 52.4\%\\
\cline{1-4}
 \small $a\text{LSTM-}VGG_{act+obj+sce}$    & (fc8) & 51.5\% & 52.8\%\\
\cline{1-4}
 \small $a\text{LSTM-}VGG_{act+obj+sce}$    &  (all)  & \textbf{52.8\%}& \textbf{54.2\%}\\
\hline
\end{tabular}
\label{tab:comparison}
\end{center}
\end{table}

Figure \ref{bestclass} shows, the top $5$ classes of ActivityNet with highest accuracy improvement with respect to vanilla LSTM-$C3D$ for $a$LSTM-$VGG_{sce}$ (fc8), $a$LSTM-$VGG_{obj}$ (fc8), and $a$LSTM-$VGG_{act}$ (fc8) models respectively. For {\em object} a number of classes (\eg, {\em pumping gas} and {\em installing car stereo}) that contain easily detectable objects, like a {\em car}, are most significantly improved (by over 30\%).

\vspace{-0.1in}
\subsection{Conclusions}
\vspace{-0.05in}
We have presented $a$LSTM model for action recognition. With this model, we are able to attent on and highlight most relevant frames and 
perform fine-grained action 
classification simultaneously. We have shown that by attending on features extracted from complementary trained VGG19 models for scene, object, and action (contextual data), we get a significant gain compared to $a$LSTM on $C3D$. Furthermore, our model variations that attend on $object$, $scene$, and $action$ are complementary which allows their combination to have the best performance. 

{\small
\bibliographystyle{ieee}
\bibliography{egbib}
}

\clearpage
\appendix

\section{Supplementary Results}\label{App:AppendixA}

Table \ref{tab:c3d1} shows average accuracy (AC), mean average precision (mAP) for Activitynet test data. In table \ref{tab:c3d1}, one could observe that our $a$LSTM models attending on $VGG_{obj}$, $VGG_{sce}$, and $VGG_{act}$ features perform better than vanilla LSTM models for the corresponding $VGG$ features; comparison to LSTM model with C3D features is in the main paper. 

We also use Hollywood2 dataset in our experiments. Hollywood2 (HOHA2) \cite{marszalek09} is a human action dataset with $12$ classes. HOHA2 includes 1707 video clips from Hollywood movies. Dataset is trimmed, but number of samples per class is not fixed and varies between 24 and 135. HOHA2 consists of 823 training samples and 884 test samples, and each clip can have multiple labels. The much smaller training set makes training LSTM-based models that much more challenging, as is also witnessed in \cite{sharma2015attention}. For HOHA2, we use multilabel classification loss
(similar to \cite{Serena972}). We initialize LSTM parameters with the pre-trained LSTM model trained for ActivityNet. We then fine-tune the model for $2,500$ iterations on minibatches of size $100$. For first $1,000$ iterations, we use fixed learning rate of $0.001$ and momentum of $0.6$; after that, learning rate is decreased every iterations by a fixed ratio.  Similar to ActivityNet model training, we also perform an on-the-fly temporal cropping of training videos. For evaluation, we report mean Average Precision (mAP) over all $12$ classes.\\
Figure \ref{fig:figVar} shows two possible variants of integrating attention with LSTM for action classification. In the first variant, Figure \ref{fig:figVar} (a), attention scores are computed at every time step $t$ and attended data is distributed into LSTM state $S_{t-1}$ to compute state $S_{t}$. Also, class scores $y_t$ are computed at every time step and final score $y$ is the average of all $y_t$s. This is the model we focused on in the approach section. 
The second variant, Figure \ref{fig:figVar} (b), has a somewhat different architecture. In particular, attention score is only computed and integrated into the last LSTM state $S_{N}$. 
The classification score $y$ is then computed directly (once) at the end. We compare both architectures, using identical setup, on HOHA2. As can be seen from Table \ref{tab:c3d} ($a$LSTM-$VGG_{obj}$ (fc8) vs. $a$LSTM-$VGG_{obj}$ (fc8)(b)), variant (b) has $2.2$ percentage points lower mAP compared to model (a).
We also evaluated significance of attended-data size, comparing $T=12$ and $T=24$. We only observed a slight improvement of $0.2$ mAP for HOHA2 using $T=24$. 
Based on these observations, we use architecture (a) with $T=12$ from now on.

\begin{table}[!t]
\begin{center}
\caption{ActivityNet: Comparitive results of vanilla LSTM and our attention-based LSTM.}
\begin{tabular}{|lr|c|c|c|c|c|c|c|}
\hline
\multicolumn{4}{|c|}{{\bf Model:} LSTM} \\
\multicolumn{2}{|c|}{Features} & AC & mAP\\
\hline
LSTM-$VGG_{obj}$ (fc6) & & 42.2\% & 47.6\% \\
\hline
$a\text{LSTM-}VGG_{obj}$ (fc6)& & {\bf 45.6\%} & {\bf 51.2\%}\\
\hline
LSTM-$VGG_{sce}$ (fc6)&  & 39.0\% & 44.4\% \\
\hline
$a\text{LSTM-}VGG_{sce}$ (fc6)& & {\bf 44.0\%} & {\bf 49.5\%}\\
\hline
LSTM-$VGG_{act}$ (fc6)&  & 45.9\% & 51.1\%\\
\hline
$a\text{LSTM-}VGG_{act}$ (fc6)& & {\bf 47.0\%} & {\bf 52.4\%} \\
\hline
\end{tabular}

\label{tab:c3d1}
\end{center}
\end{table}



Table \ref{tab:c3d} shows mean average precision over all classes (mAP) for Hollywood2 (HOHA2) test data. 
$C3D (fc8)$ shows results for 3D CNN model softmax classification directly. LSTM-$C3D$ (FC6) is vanilla LSTM model using $C3D$ model fc6 layer representation as input-data. $a$LSTM stands for our attention-based LSTM model. We have evaluated our model for different variations of attended-data: 1) $a\text{LSTM}-VGG_{obj}$, attends on objects ($VGG_{obj}$), 2) $a\text{LSTM}-VGG_{sce}$, attends on scenes ($VGG_{sce}$), and 
3) $a\text{LSTM}-VGG_{act}$ attends on actions ($VGG_{act}$).
For each of these three models we tried both fc6 and fc8 layer representations.

\begin{table}[!t]
\begin{center}
\caption{HOHA2: Our model action recognition results.}
\begin{tabular}{|lr|c|c|c|c|c|c|c|}
\hline
\multicolumn{3}{|c|}{{\bf Model:} CNN} \\
\multicolumn{2}{|c|}{Features} & mAP\\
\hline
$C3D$ & (fc8)   &  41.4\% \\
\hline
\hline
\multicolumn{3}{|c|}{{\bf Model:} LSTM} \\
\multicolumn{2}{|c|}{Features} & mAP \\
\hline
LSTM-$C3D$ & (fc6)  & 39.9\% \\
\hline
$a$LSTM-$C3D$ ~~~~~~~~& (fc8)  & 44.4\%\\
\hline
$a\text{LSTM-}VGG_{obj}$ & (fc6) & 43.6\%\\
\cline{1-3}
$a\text{LSTM-}VGG_{obj}$    & (fc8) & \textbf{46.3}\% \\
\cline{1-3}
 $a\text{LSTM-}VGG_{obj}$   & (fc8)(b) & 44.1\% \\
\hline
$a\text{LSTM-}VGG_{sce}$ & (fc6) & 43.6\%\\
\cline{1-3}
  $a\text{LSTM-}VGG_{sce}$  & (fc8) & 41.6\%\\
\hline
$a\text{LSTM-}VGG_{act}$ & (fc6) & 44.8\% \\
\cline{1-3}
$a\text{LSTM-}VGG_{act}$ & (fc8) & 45.3\%\\
\hline
\end{tabular}

\label{tab:c3d}
\end{center}
\end{table}

A few observations can be made based on Table \ref{tab:c3d}. First, LSTM on top of $C3D$ features has slightly lower performance for HOHA2 (according to mAP). All $a$LSTM models outperform vanilla LSTM and CNN models. Specifically, $a$LSTM on $C3D$ improves the performance by approximately $4$ percentage points for over vanilla LSTM. Further attending on (contextual) complementary trained VGG19 models for object and action improve the performance. However, attending on trained VGG19 scene model has lower performance compared to attending on $C3D$ directly. Among individual models, the best performance is achieved by $a$LSTM-$VGG_{obj}$ (fc8) which is consistent with recent results that show importance of objects \cite{Jian2015}.

Table \ref{tab:comparison} shows comparison of our model to state-of-the-art on HOHA2. For comparison, we have combined different variations of our models described above. Model combination has been implemented by averaging the softmax output scores over multiple $a$LSTM models. For example, $a$LSTM-$VGG_{act+obj}$ stands for combining $a$LSTM-$VGG_{act}$ and $a$LSTM-$VGG_{obj}$ and so on. Table \ref{tab:comparison} shows that our combined model for action and object improves the performance, however, adding scene model does not improve compared to our best individual model $a$LSTM-$VGG_{act}$ (fc8). 
Further, we compare our results to $DL-SFA$ \cite{6909732} and, more importantly, to ArXiv results of soft-attention model of \cite{sharma2015attention}, which proposes a spatial 3-layer attention-based LSTM.
Our best combined model $a$LSTM-$VGG_{act+obj}$ (all), outperforms \cite{sharma2015attention} by 11\% (or 5 percentage points) and has highest performance compared to methods that use RGB data directly; as opposed to working with combination of optical flow and RGB data \cite{LanLLHR15,Fernando_2015_CVPR}.   

\begin{table}[!t]
\begin{center}
\caption{ActivityNet: Comparative results.}
\begin{tabular}{|lr|c|c|c|}
\hline
\multicolumn{3}{|c|}{{\bf Model:} CNN} \\

\multicolumn{2}{|c|}{Features} & mAP \\ 
\hline
CNN-$C3D$ & (fc8)   & 41.4\%\\
\hline
\hline
\multicolumn{3}{|c|}{{\bf Model:} RGB data and/or Hand-Crafted Features} \\
\multicolumn{2}{|c|}{Features} & mAP\\ 
\hline
Soft attention model \cite{sharma2015attention} &  & 43.9\%\\ 
\hline
DL-SFA$ \cite{6909732}$ \cite{activitynet} &  & 48.1\% \\ 
\hline
\small $a\text{LSTM-}VGG_{act+obj}$ & (fc6) & 46.2\% \\
\cline{1-3}
 \small $a\text{LSTM-}VGG_{act+obj}$     & (fc8) & 48.5\% \\
\cline{1-3}
 \small $a\text{LSTM-}VGG_{act+obj}$    &  (all)  & \textbf{48.7\%}\\
\hline
\small $a\text{LSTM-}VGG_{act+sce}$ & (fc6) & 46.1\% \\
\cline{1-3}
 \small $a\text{LSTM-}VGG_{act+sce}$    & (fc8) & 45.2\% \\
\hline
\small $a\text{LSTM-}VGG_{act+obj+sce}$ & (fc6) & 46.6\% \\
\cline{1-3}
 \small $a\text{LSTM-}VGG_{act+obj+sce}$    & (fc8) & 47.9\% \\
\cline{1-3}
 \small $a\text{LSTM-}VGG_{act+obj+sce}$    &  (all)  & 48.1\%\\
\hline
\hline
\multicolumn{3}{|c|}{{\bf Model:} Hand-Crafted Features } \\
\multicolumn{2}{|c|}{Features} & HOHA2 (mAP) \\ 
\hline
Multi-skIp Feature \cite{LanLLHR15} ~~~~~~~~~~~~~~~&  & 68\% \\
\hline
VideoDarwin \cite{Fernando_2015_CVPR} &  & \textbf{73.7}\% \\
\hline

\end{tabular}
\label{tab:comparison}
\end{center}
\end{table}

\end{document}